# Scaling Item-to-Standard Alignment with Large Language Models: Accuracy, Limits, and Solutions

Farzan Karimi-Malekabadi[1,2], Pooya Razavi[1], Sonya Powers[1]

[1]Edmentum

[2]Department of Psychology, University of Southern California

**Author Note**

Farzan Karimi-Malekabadi https://orcid.org/0000-0002-3381-7071

We have no known conflict of interest to disclose.

Correspondence concerning this article should be addressed to Farzan Karimi-Malekabadi, Department of Psychology, University of Southern California, 3620 McClintock Ave, Los Angeles, CA, 90089. E-mail:karimima@usc.edu

**Abstract**

As educational systems evolve, ensuring that assessment items remain aligned with content standards is essential for maintaining fairness and instructional relevance. Traditional human alignment reviews are accurate but slow and labor-intensive, especially across large item banks. This study examines whether Large Language Models (LLMs) can accelerate this process without sacrificing accuracy. Using over 12,000 item–skill pairs in grades K–5, we tested three LLMs (GPT-3.5 Turbo, GPT-4o-mini, and GPT-4o) across three tasks that mirror real-world challenges: identifying misaligned items, selecting the correct skill from the full set of standards, and narrowing candidate lists prior to classification. In Study 1, GPT-4o-mini correctly identified alignment status in approximately 83–94% of cases, including subtle misalignments. In Study 2, performance remained strong in mathematics but was lower for reading, where standards are more semantically overlapping. Study 3 demonstrated that pre-filtering candidate skills substantially improved results, with the correct skill appearing among the top five suggestions more than 95% of the time. These findings suggest that LLMs, particularly when paired with candidate filtering strategies, can significantly reduce the manual burden of item review while preserving alignment accuracy. We recommend the development of hybrid pipelines that combine LLM-based screening with human review in ambiguous cases, offering a scalable solution for ongoing item validation and instructional alignment.

**Scaling Educational Content Alignment with Large Language Models: Accuracy, Limits, and Solutions**

Educational content alignment, the process of ensuring that assessment items are appropriately matched to the content standards they are intended to measure, is a foundational requirement for both instructional systems and standardized assessments (Webb, 2007; Porter, 2002). Accurate alignment supports valid inferences about student learning, instructional needs, and curricular effectiveness. It is also critical in standards-based adaptive and diagnostic systems, where item alignment metadata directly determines which items are administered to each student (van der Linden & Glas, 2010; von Davier, 2011). In these systems, items are tagged with one or more discrete standards, and the algorithm selects items based on the learner's estimated level of mastery for each target standard. When an item is misaligned (i.e., incorrectly labeled with a standard it does not actually assess), the system may select content that fails to target the intended construct, thereby undermining the accuracy of learner profiles and the validity of resulting interpretations (Leighton & Gierl, 2007; Rupp, Templin, & Henson, 2010).

Maintaining high-quality alignment at scale, however, presents significant practical challenges. As item banks expand and standards are revised, manually reviewing every item-standard pair becomes increasingly infeasible. Human expert review, while essential, is costly, time-intensive, and difficult to scale to the size of modern educational content libraries. Creating new items for each revision cycle introduces additional resource burdens. As a result, scalable tools that can assist in identifying and correcting misalignments are urgently needed, particularly for educational programs that rely on standards-aligned tagging to drive adaptive item delivery or instructional decision-making (Li, Pardos, & Ren, 2024).

Recent advances in Natural Language Processing (NLP) offer new possibilities for content alignment. Early embedding-based approaches, such as SentenceBERT (Matsuda et al., 2022) and MathBERT (Shen et al., 2021), showed promise in detecting clear mismatches but often struggled with subtle differences in meaning (Vahtola et al., 2022; Mahajan et al., 2024; Chiang et al., 2023). Large Language Models (LLMs) such as GPT-3 and GPT-4 have introduced models capable of much deeper contextual understanding and reasoning (Brown et al., 2020; Razavi & Powers, 2025; Wei et al., 2022). These models excel at classification, inference, and textual comprehension tasks, often matching or surpassing human performance on benchmarks in reading comprehension, question answering, and reasoning (Achiam et al., 2023). Their flexible, prompt-driven architecture supports both zero-shot and few-shot classification without task-specific fine-tuning, making them especially attractive for large-scale alignment tasks where organizations require scalable solutions but may not have extensive labeled training data.

Despite this promise, LLMs have not yet been rigorously evaluated on educational alignment tasks that reflect real-world assessment workflows. While prior work has assessed sentence-level similarity or content classification in isolation, it remains unclear whether LLMs can detect subtle semantic misalignments, generalize across educational subjects and grade levels, or support decision-making in content validation pipelines. From a measurement standpoint, alignment is not simply a semantic task; it is central to construct validity, score interpretation, and the operational logic of adaptive learning systems. Misalignment introduces construct-irrelevant variance, compromises construct representation, and disrupts the assumptions underlying item selection and scoring models (Mislevy, Steinberg, & Almond, 2002; Embretson & Reise, 2000).

To address these gaps, we conducted three empirical studies that evaluate the performance of GPT-based models on alignment tasks of increasing complexity, each grounded in the real operational challenges faced by educational content developers.

The first study focuses on misalignment detection. Here, the model is presented with an item and its associated content standards and asked to classify the pair as either aligned or misaligned. This binary classification mirrors a common alignment audit task: validating item-standard mappings in existing content libraries. By constructing misaligned examples of varying semantic similarity, ranging from cross-subject mismatches to subtle within-domain misalignments, we assess whether LLMs can differentiate degrees of conceptual fit between the item and the standard.

The second study extends the task to open-set standard classification. In this case, the model is given an item and asked to select the most appropriate standard from the full set of grade and subject-relevant standards. This reflects a realistic operational need: remapping legacy content to new or revised standards during policy transitions or curriculum updates. The classification challenge here is more cognitively demanding, as it requires reasoning over a large pool of closely related content standards. This task simulates use cases in which LLMs might assist human reviewers by narrowing candidate alignments or flagging inconsistencies.

The third study incorporates a retrieval-augmented strategy to improve performance on the open classification task. Prior to standard selection, we use semantic similarity scores from sentence embeddings to reduce the candidate standard set to the most relevant subset. The model then selects the best alignment from this smaller pool. This two-stage process draws on findings from the machine learning literature that narrowing the decision space improves both accuracy and interpretability (Reimers & Gurevych, 2019). From a content alignment perspective, it reflects a realistic pipeline in which retrieval and classification are modularized, allowing LLMs to function as downstream inference engines within human-in-the-loop validation systems.

Across all three studies, we evaluate GPT-3.5 Turbo and GPT-4o-mini using zero-shot, few-shot, and prompt-tuned variants. The dataset includes over 12,000 item-standard pairs from kindergarten through Grade 5 (K–5) math and reading content, comprising ground truth aligned and misaligned pairs. In addition to standard classification metrics such as accuracy, precision, recall, and F1 score, we examine model behavior by alignment type, subject, and grade level to provide a comprehensive view of LLM performance across varied alignment scenarios. The progressive introduction of newer GPT models across studies reflects the rapid development of LLMs and enables a layered understanding of how model capabilities have evolved, even within the relatively short span of this research.

Rather than proposing LLMs as replacements for expert review, we examine their utility as scalable decision-support tools. By empirically testing their performance across realistic alignment tasks, we aim to provide psychometricians, test developers, and curriculum designers with evidence-based guidance on when, where, and how LLMs can support the validity, efficiency, and transparency of educational content alignment. To provide this guidance, our research is structured around three specific research questions:

1. **RQ1:** How accurately can LLMs (GPT-3.5 Turbo and GPT-4o-mini) perform binary classification of item-skill alignment, and how does performance vary by the subtlety of the misalignment (completely, somewhat, vs. slightly misaligned)?

2. **RQ2:** How effective is GPT-4o-mini at an open-set classification task, selecting the correct skill for an item from the full set of grade- and subject-level standards, and how does this vary by subject (math vs. reading) and and grades??

3. **RQ3:** To what extent can a two-stage, retrieval-augmented approach (using sentence embeddings for pre-filtering) improve the accuracy of open-set skill classification?

## Study 1

To address our first research question (RQ1), this study evaluates whether LLMs can perform a binary classification task: detecting whether an item–standard pair is correctly aligned ("aligned") or incorrectly aligned ("misaligned"). In our research, these standards are represented by skill statements, the discrete, grade- and subject-specific learning objectives used to tag each assessment item in operational item banks. Each item is linked to exactly one skill statement, which serves as the basis for content delivery, diagnostic feedback, and alignment audits. This task reflects an essential step in operational test development pipelines, where psychometricians audit item-skill links to ensure construct coherence. We tested two LLMs in a binary classification task and evaluated how well they identify misalignments.

**Method**

*Constructing the Ground Truth Dataset*

To create a dataset of ground truth item-skill pairs, where the alignment status of each pair (aligned or misaligned) is known, we began with 3,011 aligned pairs drawn from math and reading skills spanning from grades K through 5. These items had been previously reviewed and manually validated by subject matter experts to ensure alignment between the item content and the corresponding skill statement. To create a comprehensive benchmark, we generated three types of misaligned pairs for each item. Including different levels of misalignment, ranging from overt cross-subject mismatches to subtle within-domain variations, allows us to assess whether models can detect not only clear errors but also fine-grained discrepancies. This design reflects the realities of operational alignment reviews, where many problematic cases are not obviously wrong but require nuanced interpretation to identify. For example, items may be mapped to a neighboring skill within the same domain, creating a subtle misfit, or mistakenly linked across subjects, creating a more obvious error. These three categories are:

- Completely Misaligned Item-Skill Pairs: These are created by pairing each item with a content standard from the same grade but a different subject. Of the three misalignment conditions, this condition represents the most contrast between the item content and content standards.

- Somewhat Misaligned Item-Skill Pairs: These are created by pairing each item with a content standard from the same subject and grade, but a different domain. Compared to the completely misaligned pairs, this condition presents a more subtle discrepancy between the item content and the corresponding content standards.

- Slightly Misaligned Item-Skill Pairs: These are created by pairing items with content standards from the same domain, subject, and grade level, but with a different skill name. This offers a minimal level of discrepancy between item contents and skill statements.

Each item was paired with one instance of each misalignment type, yielding 12,044 total item-skill pairs (3,011 aligned, 9,033 misaligned). Table 1 provides illustrative examples, and Table S1 in the supplemental materials details the proportions by grade and subject. Although these misalignments were synthetically generated, they mirror the types of errors that commonly arise in operational practice. For example, items are sometimes mapped to a neighboring skill within the same domain (slight misalignment) or mistakenly linked across subjects (complete misalignment). By designing the dataset in this way, we ensure the benchmark reflects real-world alignment challenges that psychometricians and content developers regularly face.

**Table 1**

*Generated Misaligned Item-Statement Pairs*

| Type of Pair | Item Description | Subject | Domain | Skill Name | Skill Statement | Alignment Criteria |
|---|---|---|---|---|---|---|
| **Aligned** | What fraction does the location of point S represent on the number line? | Mathematics | Fractions & Ratios | Understand Fractions Using a Number Line | Understand that fractions are numbers that can be located on a number line. Represent fractions 1/b or a/b on a number line where b is limited to 2, 3, 4, 6, or 8. | Same grade, subject, domain, and skill name |
| **Completely Misaligned** | What fraction does the location of point S represent on the number line? | Reading | Language and Vocabulary | Categories of Objects | Sort objects into categories. | Same grade, different subject, domain, and skill name |
| ***Somewhat Misaligned*** | What fraction does the location of point S represent on the number line? | Mathematics | Numbers & Operations | Multiply Using Place Value | Multiply a one-digit number by a two-digit multiple of 10 using understanding of place value and properties of operations. | Same grade and subject, different domain, and skill name |
| ***Slightly Misaligned*** | What fraction does the location of point S represent on the number line? | Mathematics | Fractions & Ratios | Express Whole Numbers as Fractions | Express whole numbers as fractions and recognize fractions that are equivalent to whole numbers. | Same grade, subject, and domain, different skill name |

***GPT Models***

We used two GPT models for the classification task: GPT-3.5 Turbo (version released on 2024-01-25) and GPT-4o-mini (version released on 2024-07-18). GPT-3.5 Turbo is based on the third generation of OpenAI's Generative Pre-Trained Transformer (GPT) architecture, featuring 175 billion parameters. It utilizes dense layers and self-attention mechanisms to process and generate text. The cost of using GPT-3.5 Turbo is $1.5 per 1 million tokens. GPT-4o-mini is part of the fourth generation of GPT models, designed with approximately 60 billion parameters. It incorporates improvements in model architecture, such as better parameter efficiency and enhanced self-attention mechanisms. The cost of using GPT-4o-mini is $0.15 per 1 million tokens. All interactions with the language model were conducted via the GPT API using the OpenAI library in Python. We compared these models based on their performance across various metrics for different grades and subjects.

*Evaluation Metrics*

We evaluated GPT's performance using standard binary classification metrics, treating misaligned pairs as the positive class. Metrics include accuracy, precision, recall, specificity, and F1 score (see Supplemental Material for definitions and formulas). Given our dataset's class imbalance, with three times as many misaligned cases, we prioritized the F1 score, which balances false positives and false negatives. This is especially important in alignment contexts where missing a misalignment can compromise a student's learning trajectory, while incorrectly flagging an aligned item simply removes it from use until human review (Powers, 2011; Sokolova & Lapalme, 2009).

*Preliminary Analyses*

When instructing GPT with a classification task, past research has shown that the prompting approach (i.e., zero-shot vs. few-shot) and the temperature setting can potentially influence the results (Brown et al., 2020). Before conducting the analyses on the complete pool of items, we ran some preliminary analyses with a subset of items (i.e., grade 2 items) to evaluate the impact of these characteristics on the model performance. While running the preliminary analyses, we requested GPT-4o-mini to provide a brief rationale for its classification decision, giving insight into the criteria used for its judgments. These rationales aided in refining our prompts for better communication.

**Zero-Shot vs. Few-Shot Learning.** We explored the model's performance using two approaches: zero-shot and few-shot learning. In the zero-shot setup, we crafted a prompt that directed GPT-4o-mini to decide on the alignment of an item's content with its corresponding skill statement without prior examples. We used empirically informed prompt engineering techniques such as the chain of thought (Wei et al., 2022). The "chain of thought" prompting technique encourages GPT to generate intermediate reasoning steps before reaching a conclusion, mimicking human problem-solving. Here is the zero-shot prompt:

> As an expert in educational assessment, your task is to determine if an assessment item aligns with a specific skill statement. Instructions: Think step by step about how the assessment item aligns with the skill statement. Please categorize the item as "aligned" or "misaligned". Your response should be only one word: "aligned" or "misaligned". Avoid labeling items as "misaligned" unless there is clear evidence.

In the few-shot approach, we introduced four examples to the model: three from different types of misalignments, all labeled as "misaligned", and one example that was "aligned". This setup aimed to enhance the model's ability to assess alignment accurately. Here is the few-shot prompt:

As an expert in educational assessment, determine if an assessment item aligns with a specific skill statement. Here are some examples to guide you:

[Followed by four examples]

Using the alignment criteria from the initial examples, determine if the following assessment item aligns with the given skill statement. Instructions: Think step by step about how the assessment item aligns with the skill statement. Categorize the item as "aligned" or "misaligned". Your response should be only one word: "aligned" or "misaligned". Avoid labeling items as "misaligned" unless there is clear evidence.

As demonstrated in Table 2, the two approaches performed similarly. Considering the higher costs of few-shot prompts (due to the additional tokens associated with providing four examples in the prompt), for the analysis of the full item pool, we proceeded with the zero-shot learning approach.

**Table 2**

*Results of Zero-Shot and Few-Shot Learning Prompts for Grade 2, Math, and Reading (GPT 4o-mini)*

| Grade | Subject | Learning Type | Accuracy | Precision | Recall | Specificity | F1 |
|-------|---------|---------------|----------|-----------|--------|-------------|------|
| 2 | Math | Zero-shot | 0.91 | 0.99 | 0.89 | 0.98 | **0.94** |
| | | Few-shot | 0.89 | 0.98 | 0.87 | 0.96 | **0.92** |
| 2 | Reading | Zero-shot | 0.83 | 0.96 | 0.81 | 0.90 | **0.88** |
| | | Few-shot | 0.84 | 0.92 | 0.86 | 0.78 | **0.89** |

**Temperature.** We also examined the effect of *temperature* to determine the optimal setting for our analysis. In the context of LLMs, temperature is a parameter that controls the randomness of the model's output: lower values (e.g., close to 0) make the model more deterministic and focused, often returning the most likely response, while higher values (e.g., closer to 1) increase variability and creativity by allowing less likely responses to be selected.

Table 3 shows the effect of temperature on classification outcomes. We found that a temperature setting of 1 improves performance in math, suggesting that a slightly increased variability helps the model capture a broader range of nuances in mathematical content alignment. However, the same setting does not significantly impact reading. To assess the stability of these findings, we repeated each condition five times using the GPT API. The F1 scores were highly consistent across runs, with standard deviations ranging from 0.003 to 0.007 and a maximum fluctuation of ±1 percentage point, indicating that the results are statistically stable despite the model's inherent randomness.

**Table 3**

*Results of zero-shot for grade 2, math, and reading by temperature (GPT 4o-mini)*

| Grade | Subject | Temperature | Accuracy | Precision | Recall | Specificity | F1 |
|-------|---------|-------------|----------|-----------|--------|-------------|------|
| 2 | Math | 1 | 0.91 | 0.99 | 0.89 | 0.98 | **0.94** |
| | | 0 | 0.84 | 0.98 | 0.8 | 0.95 | **0.88** |
| 2 | Reading | 1 | 0.83 | 0.96 | 0.81 | 0.9 | **0.88** |
| | | 0 | 0.84 | 0.96 | 0.82 | 0.9 | **0.88** |

## Results and Discussion

Based on preliminary analyses, we chose temperature 1 and zero-shot learning for the main analyses, favoring a balance between cost and performance. As shown in Table 4, the GPT 4o-mini model performed better than the GPT-3.5 Turbo model across all metrics. Based on the F1 metrics, GPT-4o had high-performance results ranging from 0.83 in grade 3 reading to 0.94 in grade 4 math, suggesting that this model is reasonably capable of understanding and evaluating content alignment.

**Table 4**

*Classification Results of Zero-Shot Model Across Grades, Subjects, and GPT Models*

| Grade | Subject | GPT Model | FP | FN | Accuracy | Precision | Recall | Specificity | F1 |
|---|---|---|---|---|---|---|---|---|---|
| K | Math | 4o-mini | 0.02 | 0.17 | 0.80 | 0.96 | 0.77 | 0.90 | **0.85** |
| | | 3.5 Turbo | 0.00 | 0.43 | 0.68 | 1.00 | 0.57 | 1.00 | **0.73** |
| K | Reading | 4o-mini | 0.01 | 0.17 | 0.81 | 0.98 | 0.76 | 0.95 | **0.86** |
| | | 3.5 Turbo | 0.06 | 0.24 | 0.69 | 0.89 | 0.67 | 0.75 | **0.76** |
| 1 | Math | 4o-mini | 0.02 | 0.09 | 0.88 | 0.96 | 0.87 | 0.89 | **0.91** |
| | | 3.5 Turbo | 0.02 | 0.17 | 0.80 | 0.96 | 0.76 | 0.91 | **0.85** |
| 1 | Reading | 4o-mini | 0.00 | 0.16 | 0.82 | 0.98 | 0.78 | 0.96 | **0.87** |
| | | 3.5 Turbo | 0.05 | 0.24 | 0.69 | 0.89 | 0.67 | 0.76 | **0.76** |
| 2 | Math | 4o-mini | 0.00 | 0.08 | 0.91 | 0.99 | 0.89 | 0.98 | **0.94** |
| | | 3.5 Turbo | 0.01 | 0.13 | 0.84 | 0.96 | 0.81 | 0.92 | **0.88** |
| 2 | Reading | 4o-mini | 0.02 | 0.13 | 0.83 | 0.96 | 0.81 | 0.90 | **0.88** |
| | | 3.5 Turbo | 0.08 | 0.17 | 0.73 | 0.86 | 0.76 | 0.64 | **0.81** |
| 3 | Math | 4o-mini | 0.01 | 0.09 | 0.89 | 0.97 | 0.87 | 0.93 | **0.92** |
| | | 3.5 Turbo | 0.01 | 0.23 | 0.74 | 0.96 | 0.68 | 0.93 | **0.80** |
| 3 | Reading | 4o-mini | 0.03 | 0.19 | 0.77 | 0.93 | 0.74 | 0.85 | **0.83** |
| | | 3.5 Turbo | 0.01 | 0.43 | 0.54 | 0.94 | 0.42 | 0.92 | **0.58** |
| 4 | Math | 4o-mini | 0.02 | 0.04 | 0.93 | 0.97 | 0.93 | 0.91 | **0.95** |
| | | 3.5 Turbo | 0.02 | 0.18 | 0.79 | 0.96 | 0.75 | 0.91 | **0.84** |
| 4 | Reading | 4o-mini | 0.00 | 0.19 | 0.80 | 0.99 | 0.74 | 0.98 | **0.84** |
| | | 3.5 Turbo | 0.03 | 0.37 | 0.59 | 0.91 | 0.50 | 0.86 | **0.64** |
| 5 | Math | 4o-mini | 0.02 | 0.05 | 0.92 | 0.96 | 0.92 | 0.90 | **0.94** |
| | | 3.5 Turbo | 0.01 | 0.22 | 0.76 | 098 | 0.70 | 0.95 | **0.81** |
| 5 | Reading | 4o-mini | 0.05 | 0.08 | 0.86 | 0.92 | 0.88 | 0.79 | **0.90** |

| | 3.5 Turbo | 0.01 | 0.32 | 0.66 | 0.97 | 0.56 | 0.95 | **0.71** |
|---|---|---|---|---|---|---|---|---|

*Note.* The false positive (FP) rate is the proportion of false positives out of the total number of classifications. The false negative (FN) rate is the proportion of false negatives out of the total number of classifications.

***Model Performance by Grade***

As shown in Figure 1, GPT-4o-mini had higher F1 scores in math compared to reading from Grades 1 through 5. This may be because reading tasks often require a deeper understanding of nuances, implied meanings, and the integration of complex narrative elements that are less formulaic than mathematical reasoning. The less structured nature of language in reading might lead to difficulties in consistently aligning text with specific skill statements, reflecting the model's relative weaknesses in handling ambiguities and the subtleties of language comprehension. Classification performance was generally stable, with all values exceeding 0.80 across grades and subjects.

**Figure 1**

*Comparison of F1 Scores by Grade and Subject for GPT 4o-mini (zero-shot classification)*

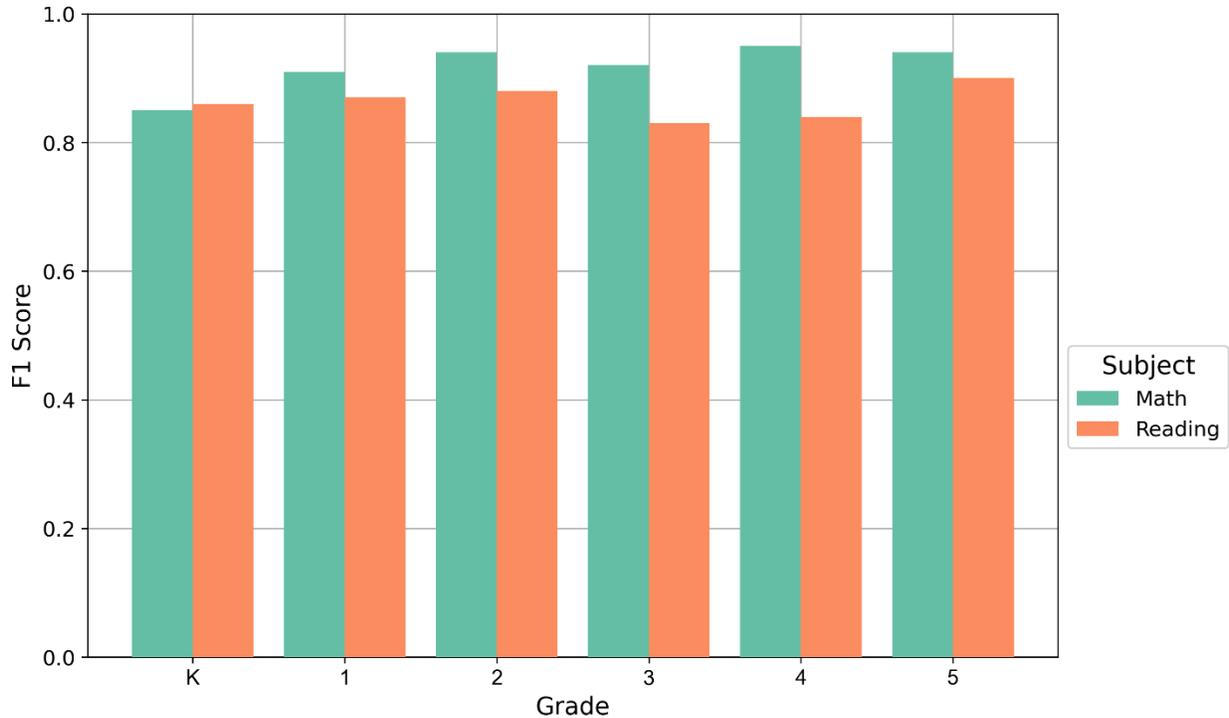

*Note*: The task is detecting whether item–skill pairs are aligned vs. misaligned. Higher F1 values indicate better model performance in distinguishing aligned from misaligned pairs.

### Model Performance by Alignment Type

We also examined performance across alignment types (aligned, completely misaligned, somewhat misaligned, and slightly misaligned). As shown in Figure 2 and Table S2 in the supplemental materials, the model was most accurate on aligned and completely misaligned pairs. Accuracy dropped notably for slightly misaligned cases, which also accounted for a large share of false negatives. Table S3 further details the distribution of false positives and false negatives by grade, subject, and misalignment type.

**Figure 2**

*Comparison of Accuracy Scores by Grade for Different Alignment Types (GPT-4o Mini) for Math (Upper Panel) and Reading (Lower Panel)*

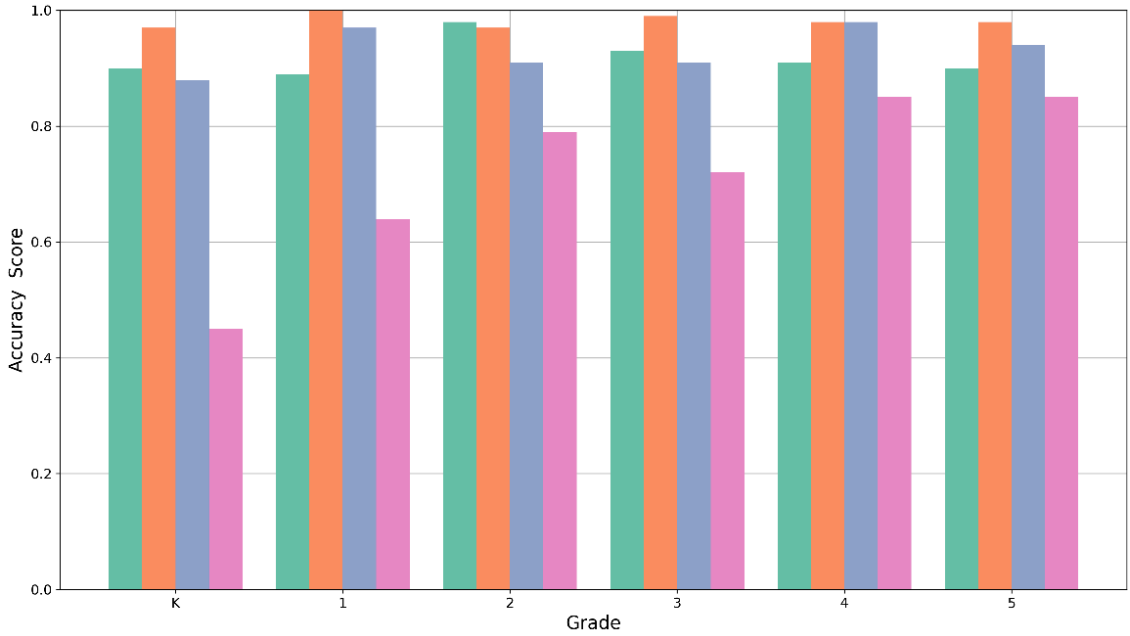

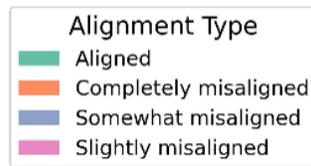

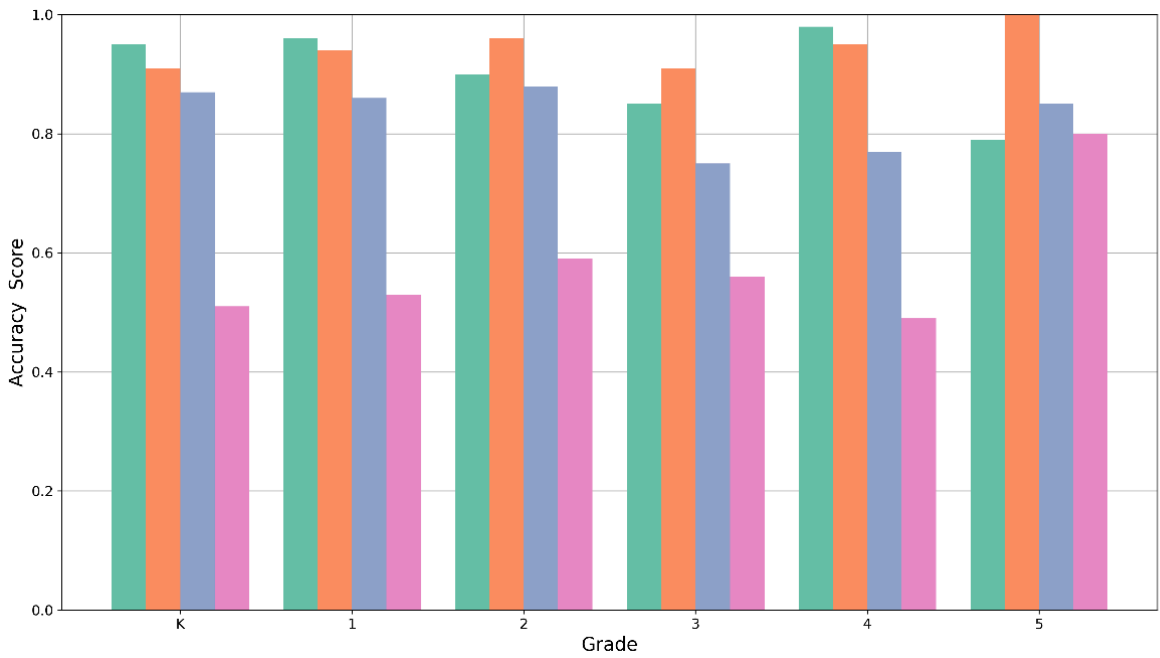

*Note*: The task is detecting whether item–skill pairs are aligned vs. misaligned.

### *Prompt Tuning to Improve Misalignment Detection*

We further examined whether prompt tuning can decrease the number of false negatives for three misalignment classes. To do so, we crafted a new prompt that aims to increase the model's sensitivity to catch misalignments. Here is the prompt that we used for this goal:

As an expert in educational assessment, your task is to determine if an assessment item aligns with a specific skill statement. Instructions: Think step by step about how the assessment item aligns with the skill statements. Please categorize the item as "aligned" or "misaligned". Your response should be only one word: "aligned" or "misaligned". Avoid labeling items as "aligned" unless there is very clear and direct evidence.

Figure 3 demonstrates that the new prompt, referred to as the second prompt, significantly improves the accuracy of detecting misaligned cases (i.e., there are fewer false negatives for all three misalignment classes). For example, the new prompt increased the minimum accuracy of the slightly misaligned class from 0.49 to 0.71 for reading and from 0.45 to 0.54 for math. In addition, the maximum accuracy for this class has risen from 0.80 to 0.87 for reading and from 0.85 to 0.92 for math.

However, this improvement comes with an important trade-off: a reduction in the accuracy of identifying aligned cases. In other words, the new prompt enhances sensitivity to misalignment but increases the likelihood of incorrectly labeling aligned items as misaligned. Choosing the appropriate prompt, therefore, requires balancing these competing priorities, maximizing the detection of misalignments while minimizing false alarms for aligned cases. The optimal balance will depend on the specific goals and constraints of the classification task.

Overall, the results, summarized in Table S4 and Figure S1 of the supplemental materials, indicate that incorporating the second prompt generally improved F1 scores across subjects and grades.

**Figure 3**

*Comparison of Accuracy Scores by Grade for Different Prompt and Alignment Types*

*(GPT-4o-mini) for Math (Upper Panel) and Reading (Lower Panel).*

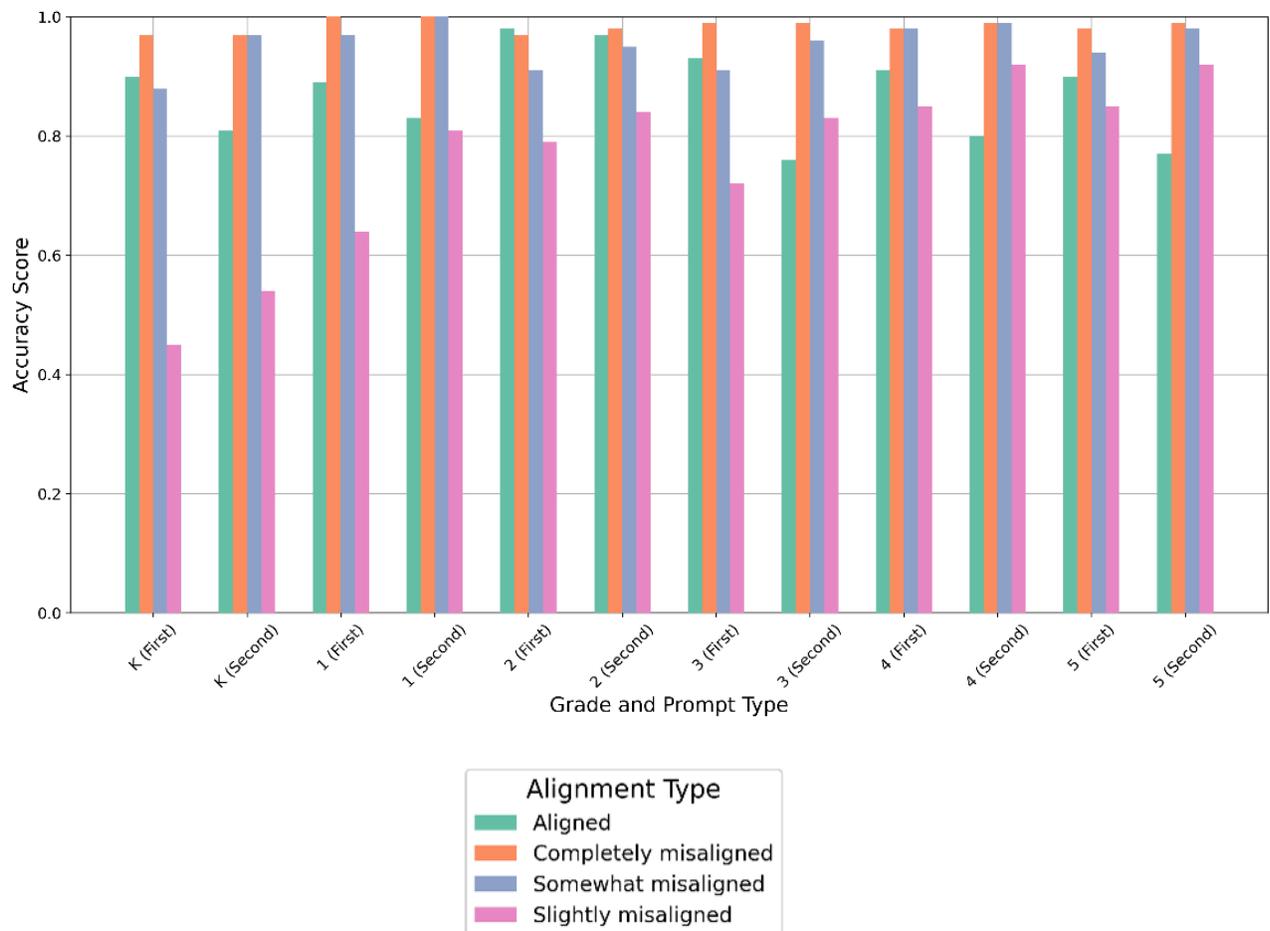

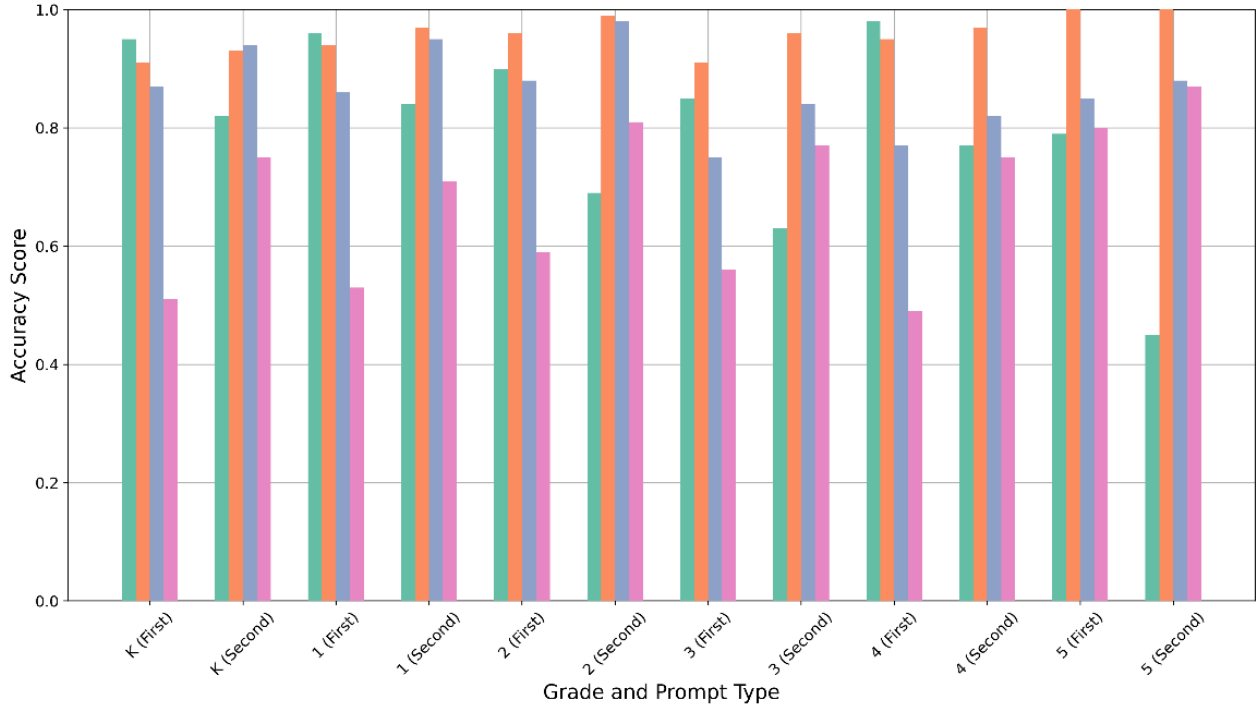

*Note*: The task is detecting whether item–skill pairs are aligned vs. misaligned. The second prompt improves sensitivity to misalignments but increases false positives for aligned cases.

**Study 2**

In Study 1, we evaluated the extent to which LLMs can effectively detect misalignments between items and skill statements. These findings suggest that LLMs are capable of evaluating conceptual coherence between assessment items and skill statements. Following these insights, a natural next question is whether LLMs can also perform a related task: not just identifying whether a given item-skill pair is valid, but selecting the correct skill from a broader set of possible candidates. This task reflects a more complex and open-ended challenge, one that better mirrors real-world scenarios where existing items must be reclassified or aligned to new standards. Study 2 addresses our second research question (RQ2) by evaluating this open classification task, specifically testing whether GPT can identify the correct skill statements for each item from the full set of grade- and subject-matched skills.

Specifically, this study evaluates whether GPT can identify the correct skill statements for each item from the full set of grade- and subject-matched skills. This open classification task reflects operational challenges in mapping legacy items to evolving standards, such as during curriculum redesigns (Leighton & Gierl, 2007; Pellegrino et al., 2001).

A key challenge in this classification task lies in the relatively large number of skills associated with each grade and subject. In our context, for instance, the number of skills per grade ranges from 12 to 36 (see Table 5). Prior research shows that classification performance decreases as the number of candidate classes increases due to increased ambiguity and decision complexity (Zheng et al., 2018), which raises important questions about the extent to which researchers and practitioners can reasonably rely on LLMs to handle this type of complex, large-scale classification task effectively.

Ultimately, the ability to classify items against a pool of candidate skills speaks directly to operational needs in educational assessment, particularly during standards revisions and curriculum redesigns. Study 2 thus represents an important step in evaluating whether GPT can meet these demands in practice.

**Table 5**

*Unique skill statements by Grade and Subject*

| Grade Level | Number of Unique Skills | |
|:---:|:---:|:---:|
| | Math | Reading |
| K | 12 | 32 |
| 1 | 15 | 30 |
| 2 | 19 | 31 |
| 3 | 26 | 33 |
| 4 | 31 | 36 |
| 5 | 22 | 33 |

**Method**

The ground truth data for this study consisted of 3,011 math and reading items (grades K through 5), each paired with the correct skill statements as determined through prior manual review by subject matter experts. We evaluated the performance of GPT-4o-mini in identifying the correct skill statements for each item in this dataset. All interactions with the language model were conducted via the GPT API using the OpenAI library in Python.

For each item, the model was provided with the item content along with all skill statements corresponding to the same grade and subject (see Table 4). For example, a grade 3 reading item was presented alongside a list of 33 relevant skill statements (i.e., all skill statements for grade 3 reading). The model was then asked to select (a) the single best-matching skill (top 1), (b) the three best-matching skills (top 3), and (c) the five best-matching skills (top 5). This was done to evaluate the model's performance under different levels of leniency, allowing us to assess not only its ability to identify the exact match but also whether the correct skill appeared among a small set of top-ranked candidates, which is a more realistic scenario for supporting human review or semi-automated alignment workflows. Each of these tasks was carried out through a separate prompt. An example prompt for the top-1 skill selection is shown below:

You are provided with an educational item and a list of skills. Your task is to determine which one skill is most likely being measured by the item. Base your decision on clear and strong evidence. WARNING: You must return ONLY the Skill ID as your response. Example Output: 520.

Accuracy for each decision type (top 1, top 3, and top 5) was calculated as the proportion of items for which the correct skill statements appeared in the model's response set, relative to the total number of items evaluated.

**Result and Discussion**

GPT-4o-mini's performance varied depending on the subject and grade. As shown in Figure 5, Top-1 accuracy for math often exceeds 70%, peaking at over 80% in Grade 2, indicating that GPT can reasonably identify the target skill among 12–31 alternatives. In contrast, detecting the correct skill statements for reading items appears to be more challenging, with Top-1 accuracy dropping as low as 39.35% in Grade 3, likely due to increased linguistic nuance and denser skill clusters.

As expected, the correct detection of the aligned skill improved markedly under Top-3 and Top-5 conditions (see Figure 5 and Table S5 in the supplemental material), with math averaging 84% and 88% and reading averaging 72% and 83%, demonstrating GPT's capacity to consistently detect the correct skill under more lenient requirements (i.e., when asked to identify the most likely candidates).

In practice, these results indicate that, when instructed to find the correctly aligned skill from a pool of skill statements, LLMs can narrow the search space effectively, a useful feature for manual auditing workflows or as a first-pass triage tool in large-scale legacy item alignment projects. However, the performance drop in reading and in mid-to-upper elementary grades highlights the limitations of unguided LLM classification when faced with subtle semantic distinctions and densely populated skill taxonomies.

**Figure 5**

*Accuracy of GPT-4o-mini in Matching Items to Skills Statements*

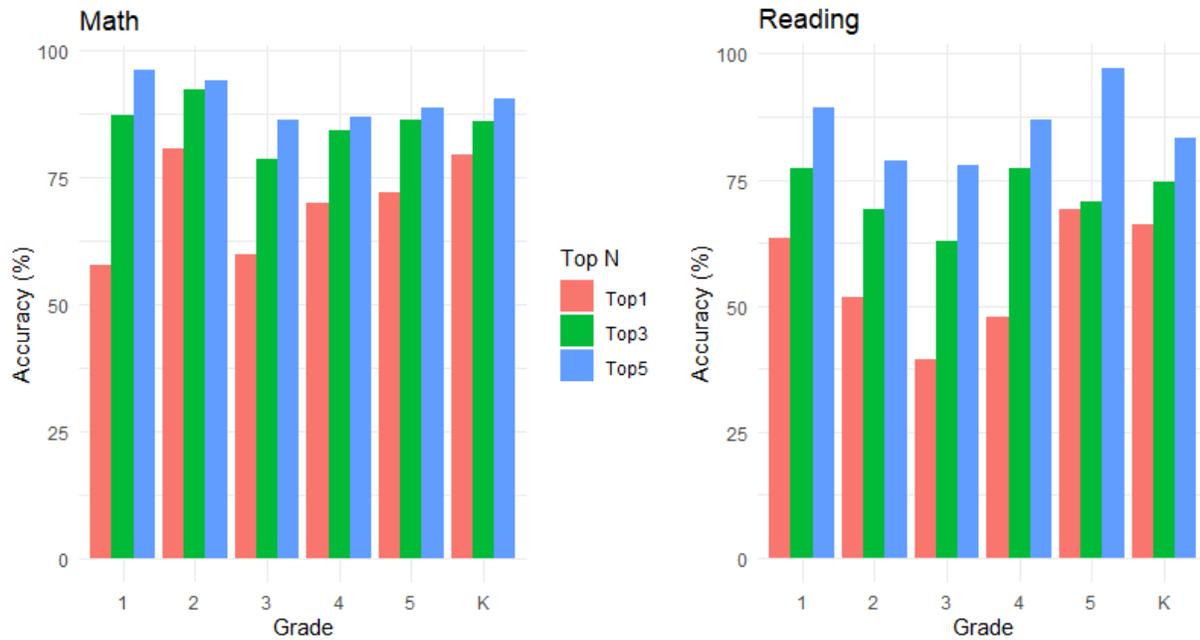

**Study 3**

While Study 2 demonstrated that GPT-4o-mini can often identify the correct skill statements from a full set of grade- and subject-matched options, its performance varied widely, particularly in reading and in grades with denser skill sets. These limitations point to the difficulty LLMs face when navigating large, semantically similar candidate pools. Study 3 addresses our third research question (RQ3) by exploring whether narrowing the search space through semantic filtering can improve model performance

**Method**

To improve GPT's performance on skill classification, in Study 3, we introduce a filtering step based on sentence embeddings. This step aims to reduce the number of candidate skills GPT has to evaluate, thereby lowering ambiguity and computational demand. We use the

all-MiniLM-L6-v2 model (Reimers & Gurevych, 2019), a lightweight, general-purpose transformer model optimized for producing high-quality sentence embeddings, to compute cosine similarity between each item (including its prompt, stem, and explanation) and all skill statements within the same grade and subject. For each item, the 15 most semantically similar skills are selected to form a filtered pool.

After this reduction step, GPT was instructed, using the same prompts as Study 2, to select (a) the single best-matching skill (top 1), (b) the three best-matching skills (top 3), and (c) the five best-matching skills (top 5) from the filtered list of skills generated in the previous step. This procedure was applied using both GPT-4o-mini and GPT-4o models.

**Results and Discussion**

Table 6 presents the overall accuracy comparison across all models and configurations. It includes the performance of the sentence embedding filter alone, GPT-4o-mini without filtering, GPT-4o-mini with filtering, and GPT-4o with filtering. GPT-4o was evaluated only in the filtered condition due to substantially higher computational cost, which made unfiltered runs infeasible at scale. The results suggest that narrowing the pool of skill candidates (using embedding-based filtering) before prompting GPT leads to consistent gains in Top 1, Top 3, and Top 5 accuracy across nearly all grade-subject combinations. For instance, GPT-4o-mini's Top 1 accuracy in math Grade K improves from 79.68% (unfiltered) to 87.5% (filtered), while GPT-4o achieves 92.18% with the same filtered input. Similar trends appear in reading, where, for example, Top 1 accuracy in Grade 1 rises from 63.32% to 67.82% with filtering using GPT-4o-mini, and further to 75.08% with GPT-4o. Notably, even the sentence similarity method on its own places the correct skill in the top 5 over 75% of the time in most math grades, establishing its effectiveness as a candidate reduction strategy (see Table S6 for additional details on the accuracy of the sentence similarity method).

These results suggest that embedding-based filtering not only reduces ambiguity in densely populated skill spaces, thereby improving classification accuracy, but also supports more efficient use of computational resources and LLM token costs by allowing high-performing models to operate on a smaller, more relevant set of candidates without sacrificing performance.

**Table 6**

*Accuracy Results for All Models and Approaches Applied in Study 3*

| Subject | Grade | Total | Top 1 | | | | Top 3 | | | | Top 5 | | | |
|---|---|---|---|---|---|---|---|---|---|---|---|---|---|---|
| | | | Sent. Emb. | 4o-mini | 4o-mini filt. | 4o filt. | Sent. Emb. | 4o-mini | 4o-mini filt. | 4o filt. | Sent. Emb. | 4o-mini | 4o-mini filt. | 4o filt. |
| Math | K | 64 | *54.69* | 79.68 | 87.50 | **92.18** | *73.44* | 85.94 | 90.62 | **96.88** | *82.81* | 90.62 | 95.31 | **98.44** |
| | 1 | 78 | *34.62* | 57.69 | 69.23 | **70.51** | *66.67* | 87.18 | 92.30 | **98.72** | *84.62* | 96.15 | 100.00 | **100.00** |
| | 2 | 103 | *50.49* | 80.58 | 80.58 | 69.90 | *71.84* | 92.23 | 91.26 | **94.17** | *80.58* | 94.17 | 93.20 | **94.17** |
| | 3 | 322 | *41.61* | 59.93 | 70.80 | **79.50** | *63.98* | 78.57 | 83.54 | **86.02** | *73.60* | 86.34 | 87.57 | **89.13** |
| | 4 | 596 | *43.62* | 70.13 | 78.85 | **80.70** | *64.60* | 84.40 | 85.81 | **87.42** | *74.33* | 87.08 | 86.82 | **87.58** |
| | 5 | 358 | *36.31* | 72.06 | 79.88 | **83.79** | *62.57* | 86.31 | 85.55 | **87.99** | *75.14* | 88.83 | 88.10 | **88.55** |
| Reading | K | 437 | *37.53* | 66.13 | 72.54 | **76.43** | *62.70* | 74.60 | 83.75 | **83.98** | *75.29* | 83.30 | 87.64 | **87.87** |
| | 1 | 289 | *35.29* | 63.32 | 67.82 | **75.08** | *52.25* | 77.16 | 80.27 | **85.81** | *63.32* | 89.27 | 86.50 | **88.93** |
| | 2 | 217 | *42.86* | 51.61 | 66.35 | **70.50** | *64.06* | 69.12 | 74.65 | **78.80** | *75.58* | 78.80 | 80.18 | **85.25** |
| | 3 | 343 | *19.24* | 39.35 | 44.02 | **49.27** | *35.28* | 62.97 | 63.74 | **64.43** | *49.56* | 77.84 | 71.05 | **73.47** |
| | 4 | 136 | *32.35* | 47.79 | 48.52 | **57.35** | *61.03* | 77.20 | 79.41 | **80.15** | *68.38* | 86.76 | 80.88 | **80.88** |
| | 5 | 68 | *8.82* | **69.11** | 61.76 | 66.17 | *42.65* | 70.58 | 80.88 | **83.82** | *58.82* | 97.05 | 83.82 | **86.76** |

*Notes.* Sent. Emb refers to Sentence Embedding, 4o-mini refers to GPT-4o-mini, 4o-mini filt. refers to GPT-4o-mini with filtering, and 4o filt. refers to GPT-4o with filtering. The 4o-mini column represents accuracy results from Study 2 (i.e., without filtering of skill statements). For each category (i.e., Top 1, Top 3, and Top 5), the lowest performance is italicized and the highest performance is in bold.

**General Discussion**

This work evaluated the feasibility of using LLMs for item-to-standard alignment by addressing three core research questions. In response to RQ1, we found that LLMs, particularly GPT-4o-mini, can effectively perform binary alignment classification, with F1 scores ranging from 0.83 to 0.94 in most cases. However, this accuracy drops significantly when models are faced with subtle, within-domain misalignments, where accuracy fell as low as 0.45 for math and 0.49 for reading. Addressing RQ2, our findings show that open-set classification from a full list of standards is a much more demanding task. Performance declined substantially, with the challenge being especially pronounced in reading, where top-1 accuracy dropped to 39.35%. Finally, in response to RQ3, we demonstrated that a retrieval-augmented approach, which uses sentence embeddings to pre-filter candidates, substantially improves classification accuracy, with GPT-4o achieving the highest scores overall. A consistent theme across all three studies was a performance gap between subjects: models consistently performed better on the structured, concrete tasks in mathematics than on the semantically nuanced and abstract tasks in reading.

Study 1 used ground-truth aligned pairs and synthetic misalignments to test GPT-4o-mini's classification accuracy across four levels of (mis)alignment (aligned, slightly misaligned, somewhat misaligned, and completely misaligned). Results revealed high accuracy for aligned cases (F1 scores from 0.83 to 0.94), validating the model's reliability under well-defined conditions. However, the model struggled with slight misalignments, where skill statements within the same domain differed subtly in intent. Accuracy in these cases dropped as low as 0.49 for reading and 0.45 for math, likely due to high semantic similarity between misaligned and aligned skill statements.

These challenges become clearer when we look at specific instances of slight misalignment. As illustrated in Figure 6, the core difficulty lies in distinguishing between near-synonymous goals, such as using "context clues" versus using "different strategies" to determine a word's meaning. Both might seem plausible responses, but only one matches the instructional intent. While current models sometimes struggle to resolve such subtleties, newer generations of LLMs are rapidly improving in capturing fine-grained distinctions. In practice, integrating a human-in-the-loop workflow can help address these edge cases by ensuring that items requiring interpretive nuance receive expert review, while the model streamlines more clear-cut decisions.

**Figure 6**

*Illustration of the Alignment Process*

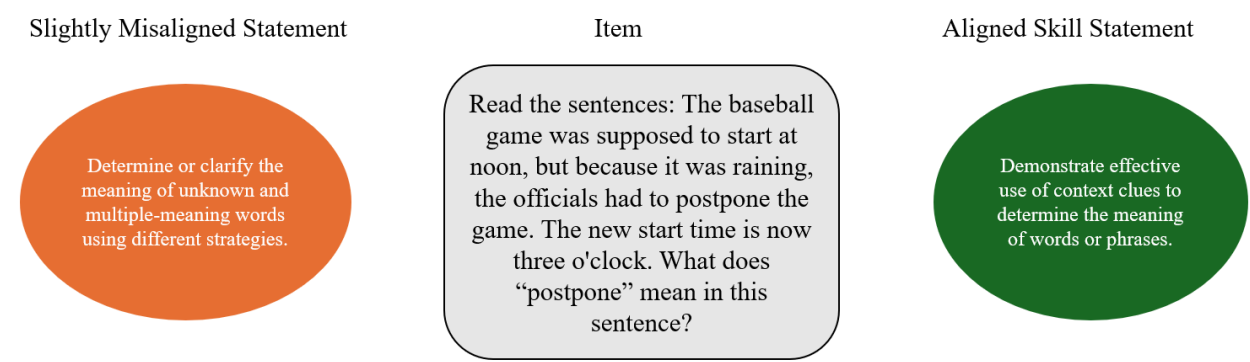

Slightly Misaligned Statement — Determine or clarify the meaning of unknown and multiple-meaning words using different strategies.

Item — Read the sentences: The baseball game was supposed to start at noon, but because it was raining, the officials had to postpone the game. The new start time is now three o'clock. What does "postpone" mean in this sentence?

Aligned Skill Statement — Demonstrate effective use of context clues to determine the meaning of words or phrases.

Prompt tuning partially addressed this issue. A revised prompt increased accuracy in the slightly misaligned class from 0.49 to 0.71 (reading) and from 0.45 to 0.54 (math), with corresponding increases in maximum accuracy. However, this improvement came at a cost—reduced precision for aligned items. This trade-off reflects a fundamental tension in prompt design: optimizing for misalignment detection may increase false positives, while optimizing for alignment detection may increase false negatives. Selecting the optimal prompt thus depends on task-specific priorities, whether to minimize the cost of missed errors or to preserve precision in high-confidence alignments.

Study 2 extended the task to an open classification setting, where GPT was asked to select the best-matching skill from the full list of grade- and subject-specific skill statements. This setting mimics real-world scenarios like curriculum redesigns or assessment content audits. LLM's accuracy declined under this more demanding setup, particularly in reading. For example, in Grade 3, top-1 accuracy was 39.35% for reading and 59.93% for math. As a whole, the Study 2 results are consistent with prior research showing that increasing the number and similarity of candidate classes degrades classification accuracy (Zheng et al., 2018). One possible reason for this degradation is that as the number of candidates increases, the model must weigh more competing semantic signals, increasing the likelihood of confusion among closely related options. Unlike humans, who might rely on deeper contextual reasoning or domain knowledge to eliminate distractors, LLMs may assign similar probabilities to semantically similar candidates, especially when distinctions are nuanced.

To address this shortcoming, Study 3 tested whether pre-filtering candidate skill statements using sentence embeddings (Reimers & Gurevych, 2019) could reduce classification complexity and improve performance. Using the all-MiniLM-L6-v2 model, we ranked each skill statement by cosine similarity to the item content and retained the top 15. This strategy yielded substantial performance gains across grades and subjects. As shown in Table 6, GPT-4o-mini's accuracy consistently increased after filtering, and GPT-4o, when combined with filtering, often achieved the highest scores overall. These findings reinforce the broader machine learning principle that narrowing the hypothesis space improves decision quality.

Despite these improvements, reading tasks continue to present significant challenges for LLMs in aligning items with the correct skill statements. One likely reason is that reading skills are often more abstract, overlapping, and context-dependent than math skills, which tend to be more concrete and procedurally defined (Snow, 2010; Sadoski, 2001). This complexity makes it harder for the model to differentiate subtle semantic distinctions, increasing the likelihood of confusion and misclassification. Psycholinguistic research documented that abstract language is harder to process than concrete language (Brysbaert et al., 2014), and embedding studies suggest that semantically similar statements form denser clusters in representational space (Reimers & Gurevych, 2019), both of which may contribute to the observed performance gap. For example, in a Grade 5 reading item that asked students to interpret the meaning of the word *tragic* within a short passage, GPT selected a skill focused on general vocabulary acquisition strategies rather than the intended skill, which emphasized using context clues to infer word meaning. While both skills were closely related and semantically similar, this example illustrates how the model can struggle to identify the most specific and instructionally appropriate match when skills are abstract and conceptually adjacent.

To further improve performance, a promising direction for future research is the development of hierarchical classification frameworks. In such an approach, the model would first classify an item into a broad domain (e.g., "Reading Comprehension" or "Phonics") and then identify the best-matching skill within that domain. This top-down narrowing aligns with theory-driven test design and may ultimately outperform bottom-up filtering based on semantic similarity, particularly in operational contexts where domain-level structures are already in place.

We also recommend integrating an ensemble approach combined with item-level confidence interval reporting. This can be achieved by running the model multiple times for each item using varied temperature settings, which introduces controlled randomness into the prediction process. The resulting distribution of predictions can then be used to calculate confidence intervals around the model's skill selection. While confidence intervals and uncertainty estimation techniques are common in NLP, for example, through modeling data and model uncertainties (Xiao & Wang, 2018), semantic entropy (Kuhn et al., 2023), and comprehensive uncertainty calibration studies in LLMs (Tao et al., 2025), applying this mechanism to quantify uncertainty at the individual decision level, that is, around a specific skill selection, is a novel application in the educational alignment context. This approach offers more than just a single prediction; it provides a measure of the model's certainty for each item. Such information can be especially valuable in practice, enabling educators and assessment designers to flag items with low confidence scores for further human review while relying on high-confidence predictions to streamline alignment workflows. Ultimately, this could support more efficient and trustworthy semi-automated item classification systems.

Another area for future research arises from our finding that aligning reading items with skill statements appears to be especially challenging for LLMs. A natural next step is to systematically examine how human experts approach this process. Do experts encounter the same challenges when mapping reading items to skill statements? Mixed-method research could provide valuable insights into the strategies, reasoning, and challenges human reviewers face (Polikoff, 2012). Such findings may help identify the underlying sources of difficulty, since reading standards often employ abstract and overlapping language (Snow & Uccelli, 2009). Moreover, if both LLMs and human experts struggle in similar ways, this would point to deeper issues in the construction of the standards themselves. In that case, alternative approaches, such as developing semantic rules or refining standards to be more concrete and less overlapping (Porter & Smithson, 2001), may be needed to improve clarity and reduce ambiguity in future assessment design.

Finally, future work can explore the development of a comprehensive diagnostic pipeline that integrates both misalignment detection (as demonstrated in Study 1) and best-skill recommendation (as explored in Studies 2 and 3). Such a system could first flag potentially misaligned item-skill pairs and then select more appropriate skill statements through open-set classification. This two-stage approach has the potential to significantly reduce the manual burden of reviewing large item banks, offering a scalable solution for aligning educational content with evolving standards.

Our results underscore that while GPT models are already effective alignment tools, their performance can be significantly enhanced through targeted task design, prompt engineering, and prefiltering through the use of NLP methods. Importantly, scale and consistency are strengths of new AI models, while their limitations, particularly in edge cases, underscore the continuing need for human oversight. Hybrid systems that combine the speed of AI with the judgment of experts will continue to be essential for ensuring content quality and alignment fidelity.

## References

Achiam, J., Adler, S., Agarwal, S., Ahmad, L., Akkaya, I., Alemán, F. L., ... & McGrew, B. (2023). GPT-4 technical report. *arXiv preprint arXiv:2303.08774.* https://arxiv.org/abs/2303.08774

Brown, T., Mann, B., Ryder, N., Subbiah, M., Kaplan, J., Dhariwal, P., ... & Amodei, D. (2020). Language models are few-shot learners. *Advances in Neural Information Processing Systems, 33,* 1877–1901.

Brysbaert, M., Warriner, A. B., & Kuperman, V. (2014). Concreteness ratings for 40,000 generally known English word lemmas. *Behavior Research Methods, 46*(3), 904–911. https://doi.org/10.3758/s13428-013-0403-5

Chiang, C. H., Chuang, Y. S., Glass, J., & Lee, H. Y. (2023). Revealing the blind spot of sentence encoder evaluation by HEROS. *arXiv preprint arXiv:2306.05083.* https://arxiv.org/abs/2306.05083

Davier, A. A. (Ed.). (2011). *Statistical models for test equating, scaling, and linking.* Springer Science+Business Media.

Embretson, S. E., & Reise, S. P. (2000). *Item response theory for psychologists* (1st ed.). Psychology Press. https://doi.org/10.4324/9781410605269

Kuhn, L., Gal, Y., & Farquhar, S. (2023). Semantic uncertainty: Linguistic invariances for uncertainty estimation in natural language generation. In *Proceedings of the International Conference on Learning Representations (ICLR).* https://arxiv.org/abs/2302.09664

Leighton, J., & Gierl, M. (Eds.). (2007). *Cognitive diagnostic assessment for education: Theory and applications.* Cambridge University Press.

Li, Z., Pardos, Z. A., & Ren, C. (2024). Aligning open educational resources to new taxonomies: How AI technologies can help and in which scenarios. *Computers & Education, 216,* 105027. https://doi.org/10.1016/j.compedu.2024.105027

Mahajan, Y., Bansal, N., Blanco, E., & Karmaker, S. (2024). ALIGN-SIM: A task-free test bed for evaluating and interpreting sentence embeddings through semantic similarity alignment. In *Findings of the Association for Computational Linguistics: EMNLP 2024* (pp. 7393–7428). https://doi.org/10.18653/v1/2024.findings-emnlp.492

Matsuda, N., Wood, J., Shrivastava, R., Shimmei, M., & Bier, N. (2022). Latent skill mining and labeling from courseware content. *Journal of Educational Data Mining, 14*(2), 22–45.

Mislevy, R. J., Steinberg, L. S., & Almond, R. G. (2002). Design and analysis in task-based language assessment. *Language Testing, 19*(4), 477–496. https://doi.org/10.1191/0265532202lt239oa

Pellegrino, J. W., Chudowsky, N., & Glaser, R. (Eds.). (2001). *Knowing what students know: The science and design of educational assessment.* National Academy Press. https://doi.org/10.17226/10019

Polikoff, M. S. (2012). Instructional alignment under No Child Left Behind. *American Journal of Education, 118*(3), 341–368. https://doi.org/10.1086/664773

Porter, A. C., & Smithson, J. L. (2001). *Defining, developing, and using curriculum indicators.* Consortium for Policy Research in Education.

Porter, A. C. (2002). Measuring the content of instruction: Uses in research and practice. *Educational Researcher, 31*(7), 3–14. https://doi.org/10.3102/0013189X031007003

Powers, D. M. W. (2011). Evaluation: From precision, recall and F-measure to ROC, informedness, markedness, and correlation. *Journal of Machine Learning Technologies, 2*(1), 37–63.

Razavi, P., & Powers, S. J. (2025). Estimating item difficulty using Large Language Models and tree-based machine learning algorithms. *arXiv preprint arXiv:2504.08804.* https://arxiv.org/abs/2504.08804

Reimers, N., & Gurevych, I. (2019). Sentence-BERT: Sentence embeddings using Siamese BERT-networks. In *Proceedings of the 2019 Conference on Empirical Methods in Natural Language Processing and the 9th International Joint Conference on Natural Language Processing (EMNLP-IJCNLP)* (pp. 3982–3992). Association for Computational Linguistics. https://doi.org/10.18653/v1/D19-1410

Rupp, A. A., Templin, J., & Henson, R. A. (2010). *Diagnostic measurement: Theory, methods, and applications.* The Guilford Press.

Sadoski, M. (2001). Resolving the effects of concreteness on interest, comprehension, and learning important ideas from text. *Educational Psychology Review, 13*(3), 263–281. https://doi.org/10.1023/A:101667582293

Shen, J. T., Yamashita, M., Prihar, E., Heffernan, N., Wu, X., Graff, B., & Lee, D. (2021, December). MathBERT: A pre-trained language model for general NLP tasks in mathematics education. In *NeurIPS 2021 Math AI for Education Workshop.*

Snow, C. E., & Uccelli, P. (2009). The challenge of academic language. In D. R. Olson & N. Torrance (Eds.), *The Cambridge handbook of literacy* (pp. 112–133). Cambridge

Snow, C. E. (2010). Academic language and the challenge of reading for learning about science. *Science, 328*(5977), 450–452. https://doi.org/10.1126/science.1182597

Sokolova, M., & Lapalme, G. (2009). A systematic analysis of performance measures for classification tasks. *Information Processing & Management, 45*(4), 427–437. https://doi.org/10.1016/j.ipm.2009.03.002

Tao, L., Yeh, Y.-F., Dong, M., Huang, T., Torr, P., & Xu, C. (2025). Revisiting uncertainty estimation and calibration of large language models. *arXiv preprint arXiv:2505.23854.* https://arxiv.org/abs/2505.23854

University Press. https://doi.org/10.1017/CBO9780511609664.008

Vahtola, T., Creutz, M., & Tiedemann, J. (2022). It is not easy to detect paraphrases: Analysing semantic similarity with antonyms and negation using the new SemAntoNeg benchmark. In *Proceedings of the Fifth BlackboxNLP Workshop on Analyzing and Interpreting Neural Networks for NLP* (pp. 249–262). Association for Computational Linguistics. https://doi.org/10.18653/v1/2022.blackboxnlp-1.22

Van der Linden, W. J., & Glas, C. A. (Eds.). (2010). *Elements of adaptive testing* (Vol. 10). Springer.

Webb, N. L. (2007). Issues related to judging the alignment of curriculum standards and assessments. *Applied Measurement in Education, 20*(1), 7–25. https://doi.org/10.1080/08957340709336746

Wei, J., Wang, X., Schuurmans, D., Bosma, M., Xia, F., Chi, E., ... & Zhou, D. (2022). Chain-of-thought prompting elicits reasoning in large language models. *Advances in Neural Information Processing Systems, 35,* 24824–24837.

Xiao, Y., & Wang, W. Y. (2018). Quantifying uncertainties in natural language processing tasks. *arXiv preprint arXiv:1811.07253.* https://arxiv.org/abs/1811.07253

Zheng, C., Achanta, R., & Benjamini, Y. (2018). Extrapolating expected accuracies for large multi-class problems. *Journal of Machine Learning Research, 19*(65), 1–30. https://doi.org/10.5555/3291125.3309627

**Supplemental Materials**

<div align="center">

**Study 1**

</div>

**Table S1**

*Proportions of Ground-Truth Items Relative to Total Items by Subject and Grade.*

| Subject | Grade | Proportion |
| --- | --- | --- |
| Mathematics | K | 0.2 |
| Mathematics | 1 | 0.44 |
| Mathematics | 2 | 0.18 |
| Mathematics | 3 | 0.38 |
| Mathematics | 4 | 0.59 |
| Mathematics | 5 | 0.53 |
| Reading | K | 0.55 |
| Reading | 1 | 0.38 |
| Reading | 2 | 0.26 |
| Reading | 3 | 0.44 |
| Reading | 4 | 0.19 |
| Reading | 5 | 0.11 |

**Table S2**

*Generated Misaligned Item-Statement Pairs*

| Type of Pair | Item Description | Subject | Domain | Skill Name | Skill Statement | Alignment Criteria |
|---|---|---|---|---|---|---|
| **Aligned** | What fraction does the location of point S represent on the number line? | Mathematics | Fractions & Ratios | Understand Fractions Using a Number Line | Understand that fractions are numbers that can be located on a number line. Represent fractions 1/b or a/b on a number line where b is limited to 2, 3, 4, 6, or 8. | Same grade, subject, domain, and skill name |

| **Completely Misaligned** | What fraction does the location of point S represent on the number line? | Reading | Language and Vocabulary | Categories of Objects | Sort objects into categories. | Same grade, different subject, domain, and skill name |
| **Somewhat Misaligned** | What fraction does the location of point S represent on the number line? | Mathematics | Numbers & Operations | Multiply Using Place Value | Multiply a one-digit number by a two-digit multiple of 10 using understanding of place value and properties of operations. | Same grade and subject, different domain, and skill name |
| **Slightly Misaligned** | What fraction does the location of point S represent on the number line? | Mathematics | Fractions & Ratios | Express Whole Numbers as Fractions | Express whole numbers as fractions and recognize fractions that are equivalent to whole numbers. | Same grade, subject, domain, different skill name |

## Evaluation Metrics

We evaluated GPT's performance using standard classification metrics. Because our main purpose here is to detect misalignments, in our results, "positive" refers to "misaligned," and "negative" refers to "aligned." A true positive (TP) indicates an item correctly predicted as "misaligned," while a true negative (TN) indicates an item correctly predicted as "aligned." A false positive (FP) refers to an item incorrectly identified as "misaligned," and a false negative (FN) refers to an item wrongly identified as "aligned."

Additionally, we report four metrics that assess the classification performance in different ways. Accuracy represents the proportion of correct predictions (both "aligned" and "misaligned"):

$$Accuracy \ = \ \frac{TP+TN}{TP+TN+FP+FN}$$

Precision measures the proportion of correctly "misaligned" predictions out of all "misaligned" predictions:

$$Precision \ = \ \frac{TP}{TP+FP}$$

Recall (or sensitivity) represents the proportion of correct "misaligned" predictions out of all actual (both correctly identified and missed) "misaligned" items-statement pairs:

$$Recall \ = \ \frac{TP}{TP+FN}$$

Specificity measures the proportion of correct "aligned" predictions out of all actual (both correctly identified and missed) "aligned" items:

$$Specificity \ = \ \frac{TN}{TN+FP}$$

F1 Score is the harmonic mean of Precision and Recall for the "misaligned" category:

$$F1 \ Score \ = \ 2 \times \frac{Precision \times Recall}{Precision+ \ Recall}$$

***Preliminary Analyses***

**Zero-Shot vs. Few-Shot Learning.** We explored the model's performance using two approaches: zero-shot and few-shot learning. In the zero-shot setup, we crafted a prompt that directed GPT-4o-mini to decide on the alignment of an item's content with its corresponding skill statement without prior examples. We used empirically informed prompt engineering techniques such as the chain of thought (Wei et al., 2022). The "chain of thought" prompting technique encourages GPT to generate intermediate reasoning steps before reaching a conclusion, mimicking human problem-solving. Here is the zero-shot prompt:

> As an expert in educational assessment, your task is to determine if an assessment item aligns with a specific skill statement. Instructions: Think step by step about how the assessment item aligns with the skill statement. Please categorize the item as "aligned" or "misaligned". Your response should be only one word: "aligned" or "misaligned". Avoid labeling items as "misaligned" unless there is clear evidence.

In the few-shot approach, we introduced four examples to the model: three from different types of misalignments, all labeled as "misaligned", and one example that was "aligned". This setup aimed to enhance the model's ability to assess alignment accurately. Here is the few-shot prompt:

As an expert in educational assessment, determine if an assessment item aligns with a specific skill statement. Here are some examples to guide you:

[Followed by four examples]

Using the alignment criteria from the initial examples, determine if the following assessment item aligns with the given skill statement. Instructions: Think step by step about how the assessment item aligns with the skill statement. Categorize the item as "aligned" or "misaligned". Your response should be only one word: "aligned" or "misaligned". Avoid labeling items as "misaligned" unless there is clear evidence.

As demonstrated in Table S3, the two approaches performed similarly. Considering the higher costs of few-shot prompts (due to the additional tokens associated with providing four examples in the prompt), for the analysis of the full item pool, we proceeded with the zero-shot learning approach.

**Table S3**

*Results of Zero-Shot and Few-Shot Learning Prompts for Grade 2, Math, and Reading (GPT 4o-mini)*

| Grade | Subject | Learning Type | Accuracy | Precision | Recall | Specificity | F1 |
|-------|---------|---------------|----------|-----------|--------|-------------|------|
| **2** | **Math** | Zero-shot | 0.91 | 0.99 | 0.89 | 0.98 | **0.94** |
| | | Few-shot | 0.89 | 0.98 | 0.87 | 0.96 | **0.92** |
| **2** | **Reading** | Zero-shot | 0.83 | 0.96 | 0.81 | 0.90 | **0.88** |
| | | Few-shot | 0.84 | 0.92 | 0.86 | 0.78 | **0.89** |

**Temperature.** We also examined the effect of *temperature* to determine the optimal setting for our analysis. In the context of LLMs, temperature is a parameter that controls the randomness of the model's output: lower values (e.g., close to 0) make the model more deterministic and focused, often returning the most likely response, while higher values (e.g., closer to 1) increase variability and creativity by allowing less likely responses to be selected.

Table S4 shows the effect of temperature on classification outcomes. We found that a temperature setting of 1 improves performance in math, suggesting that a slightly increased variability helps the model capture a broader range of nuances in mathematical content alignment. However, the same setting does not significantly impact reading. To assess the stability of these findings, we repeated each condition five times using the GPT API. The F1 scores were highly consistent across runs, with standard deviations ranging from 0.003 to 0.007 and a maximum fluctuation of ±1 percentage point, indicating that the results are statistically stable despite the model's inherent randomness.

**Table S4**

*Results of zero-shot for grade 2, math, and reading by temperature (GPT 4o-mini)*

| Grade | Subject | Temperature | Accuracy | Precision | Recall | Specificity | F1 |
|---|---|---|---|---|---|---|---|
| **2** | **Math** | 1 | 0.91 | 0.99 | 0.89 | 0.98 | **0.94** |
| | | 0 | 0.84 | 0.98 | 0.8 | 0.95 | **0.88** |

| | | 1 | 0.83 | 0.96 | 0.81 | 0.9 | **0.88** |
|---|---|---|---|---|---|---|---|
| 2 | Reading | | | | | | |
| | | 0 | 0.84 | 0.96 | 0.82 | 0.9 | **0.88** |

## Table S5

*Classification Results of Zero-Shot Model Across Grades, Subjects, and GPT Models*

| Grade | Subject | GPT Model | FP | FN | Accuracy | Precision | Recall | Specificity | F1 |
|---|---|---|---|---|---|---|---|---|---|
| K | Math | 4o-mini | 0.02 | 0.17 | 0.80 | 0.96 | 0.77 | 0.90 | **0.85** |
| | | 3.5 Turbo | 0.00 | 0.43 | 0.68 | 1.00 | 0.57 | 1.00 | **0.73** |
| K | Reading | 4o-mini | 0.01 | 0.17 | 0.81 | 0.98 | 0.76 | 0.95 | **0.86** |
| | | 3.5 Turbo | 0.06 | 0.24 | 0.69 | 0.89 | 0.67 | 0.75 | **0.76** |
| 1 | Math | 4o-mini | 0.02 | 0.09 | 0.88 | 0.96 | 0.87 | 0.89 | **0.91** |
| | | 3.5 Turbo | 0.02 | 0.17 | 0.80 | 0.96 | 0.76 | 0.91 | **0.85** |
| 1 | Reading | 4o-mini | 0.00 | 0.16 | 0.82 | 0.98 | 0.78 | 0.96 | **0.87** |
| | | 3.5 Turbo | 0.05 | 0.24 | 0.69 | 0.89 | 0.67 | 0.76 | **0.76** |
| 2 | Math | 4o-mini | 0.00 | 0.08 | 0.91 | 0.99 | 0.89 | 0.98 | **0.94** |
| | | 3.5 Turbo | 0.01 | 0.13 | 0.84 | 0.96 | 0.81 | 0.92 | **0.88** |
| 2 | Reading | 4o-mini | 0.02 | 0.13 | 0.83 | 0.96 | 0.81 | 0.90 | **0.88** |
| | | 3.5 Turbo | 0.08 | 0.17 | 0.73 | 0.86 | 0.76 | 0.64 | **0.81** |
| 3 | Math | 4o-mini | 0.01 | 0.09 | 0.89 | 0.97 | 0.87 | 0.93 | **0.92** |
| | | 3.5 Turbo | 0.01 | 0.23 | 0.74 | 0.96 | 0.68 | 0.93 | **0.80** |
| 3 | Reading | 4o-mini | 0.03 | 0.19 | 0.77 | 0.93 | 0.74 | 0.85 | **0.83** |
| | | 3.5 Turbo | 0.01 | 0.43 | 0.54 | 0.94 | 0.42 | 0.92 | **0.58** |
| 4 | Math | 4o-mini | 0.02 | 0.04 | 0.93 | 0.97 | 0.93 | 0.91 | **0.95** |
| | | 3.5 Turbo | 0.02 | 0.18 | 0.79 | 0.96 | 0.75 | 0.91 | **0.84** |
| 4 | Reading | 4o-mini | 0.00 | 0.19 | 0.80 | 0.99 | 0.74 | 0.98 | **0.84** |
| | | 3.5 Turbo | 0.03 | 0.37 | 0.59 | 0.91 | 0.50 | 0.86 | **0.64** |
| 5 | Math | 4o-mini | 0.02 | 0.05 | 0.92 | 0.96 | 0.92 | 0.90 | **0.94** |
| | | 3.5 Turbo | 0.01 | 0.22 | 0.76 | 098 | 0.70 | 0.95 | **0.81** |
| 5 | Reading | 4o-mini | 0.05 | 0.08 | 0.86 | 0.92 | 0.88 | 0.79 | **0.90** |
| | | 3.5 Turbo | 0.01 | 0.32 | 0.66 | 0.97 | 0.56 | 0.95 | **0.71** |

*Note.* The false positive (FP) rate is the proportion of false positives out of the total number of classifications. The false negative (FN) rate is the proportion of false negatives out of the total number of classifications.

**Model Performance by Grade and Alignment Type**

**Table S6**

*Accuracy for Different Types of Alignment (Zero-Shot)*

| Grade | Subject | GPT Model | Aligned | Completely misaligned | Somewhat misaligned | Slightly misaligned |
|-------|---------|-----------|---------|-----------------------|---------------------|---------------------|
| **K** | **Math** | 4o-mini | 0.90 | 0.97 | 0.88 | 0.45 |
|       |          | 3.5 Turbo | 1 | 0.95 | 0.54 | 0.22 |
| **K** | **Reading** | 4o-mini | 0.95 | 0.91 | 0.87 | 0.51 |
|       |          | 3.5 Turbo | 0.75 | 0.79 | 0.72 | 0.49 |
| **1** | **Math** | 4o-mini | 0.89 | 1.00 | 0.97 | 0.64 |
|       |          | 3.5 Turbo | 0.91 | 0.95 | 0.83 | 0.50 |
| **1** | **Reading** | 4o-mini | 0.96 | 0.94 | 0.86 | 0.53 |
|       |          | 3.5 Turbo | 0.76 | 0.82 | 0.69 | 0.49 |
| **2** | **Math** | GPT 4o-mini | 0.98 | 0.97 | 0.91 | 0.79 |
|       |          | 3.5 Turbo | 0.92 | 0.91 | 0.84 | 0.69 |
| **2** | **Reading** | 4o-mini | 0.90 | 0.96 | 0.88 | 0.59 |
| | | 3.5 Turbo | 0.64 | 0.88 | 0.64 | 0.59 |
| **3** | **Math** | 4o-mini | 0.93 | 0.99 | 0.91 | 0.72 |
| | | 3.5 Turbo | 0.93 | 0.91 | 0.67 | 0.45 |
| **3** | **Reading** | 4o-mini | 0.85 | 0.91 | 0.75 | 0.56 |
| | | 3.5 Turbo | 0.92 | 0.70 | 0.38 | 0.17 |
| **4** | **Math** | 4o-mini | 0.91 | 0.98 | 0.98 | 0.85 |
| | | GPT-3.5 Turbo | 0.91 | 0.92 | 0.81 | 0.53 |
| **4** | **Reading** | GPT 4o-mini | 0.98 | 0.95 | 0.77 | 0.49 |
| | | GPT-3.5 Turbo | 0.86 | 0.86 | 0.44 | 0.29 |
| **5** | **Math** | GPT 4o-mini | 0.90 | 0.98 | 0.94 | 0.85 |
| | | GPT-3.5 Turbo | 0.95 | 0.92 | 0.68 | 0.48 |
| **5** | **Reading** | GPT 4o-mini | 0.79 | 1.00 | 0.85 | 0.80 |
| | | GPT-3.5 Turbo | 0.95 | 0.82 | 0.45 | 0.41 |

**Table S7**

*The Number of False Positives and False Negatives for Different Types of Alignment (Zero-Shot)*

| Grade | Subject | GPT Model | FP (A) | FN (CM) | FN (SOM) | FN (SLM) | Total Predictions |
|-------|---------|-----------|--------|---------|----------|----------|-------------------|
| K | Math | 4o mini | 4 | 1 | 5 | 24 | 176 |
| | | 3.5 Turbo | 0 | 2 | 20 | 34 | 176 |
| K | Reading | 4o mini | 20 | 36 | 53 | 209 | 1724 |
| | | 3.5 Turbo | 104 | 87 | 120 | 218 | 1724 |
| 1 | Math | 4o mini | 0 | 5 | 1 | 17 | 192 |
| | | 3.5 Turbo | 4 | 2 | 8 | 24 | 192 |
| 1 | Reading | 4o mini | 9 | 15 | 38 | 135 | 1156 |
| | | 3.5 Turbo | 68 | 50 | 87 | 146 | 1156 |
| 2 | Math | 4o mini | 2 | 3 | 9 | 21 | 412 |
| | | 3.5 Turbo | 8 | 9 | 16 | 31 | 412 |
| 2 | Reading | 4o mini | 21 | 7 | 25 | 87 | 868 |
| | | 3.5 Turbo | 76 | 26 | 42 | 87 | 868 |
| 3 | Math | 4o mini | 21 | 1 | 26 | 85 | 1240 |
| | | 3.5 Turbo | 21 | 26 | 101 | 169 | 1240 |
| 3 | Reading | 4o mini | 50 | 28 | 85 | 147 | 1360 |
| | | 3.5 Turbo | 27 | 99 | 210 | 280 | 1360 |

| | | | A | CM | SOM | SLM | |
|---|---|---|---|---|---|---|---|
| 4 | Math | 4o mini | 51 | 9 | 10 | 88 | 2352 |
| | | 3.5 Turbo | 49 | 46 | 107 | 274 | 2352 |
| 4 | Reading | 4o mini | 2 | 6 | 29 | 65 | 516 |
| | | 3.5 Turbo | 17 | 31 | 71 | 91 | 516 |
| 5 | Math | 4o mini | 34 | 4 | 18 | 52 | 1396 |
| | | 3.5 Turbo | 14 | 25 | 109 | 180 | 1396 |
| 5 | Reading | 4o mini | 13 | 0 | 9 | 12 | 248 |
| | | 3.5 Turbo | 3 | 11 | 34 | 36 | 248 |

*Note.* A, CM, SOM, and SLM refer to aligned, completely misaligned, somewhat misaligned, and slightly misaligned, respectively.

### *Prompt Tuning to Improve Misalignment Detection*

We further examined whether prompt tuning can decrease the number of false negatives for three misalignment classes. To do so, we crafted an alternative prompt designed to increase the model's sensitivity to misalignments. Here is the prompt that we used for this goal:

As an expert in educational assessment, your task is to determine if an assessment item aligns with a specific skill statement. Instructions: Think step by step about how the assessment item aligns with the skill statements. Please categorize the item as "aligned" or "misaligned". Your response should be only one word: "aligned" or "misaligned". Avoid labeling items as "aligned" unless there is very clear and direct evidence.

Figure S1 demonstrates that the new prompt, referred to as the second prompt, significantly improves the accuracy of detecting misaligned cases (i.e., there are fewer false negatives for all three misalignment classes). For example, the new prompt increased the minimum accuracy of the slightly misaligned class from 0.49 to 0.71 for reading and from 0.45 to 0.54 for math. In addition, the maximum accuracy for this class has risen from 0.80 to 0.87 for reading and from 0.85 to 0.92 for math.

However, this improvement comes with an important trade-off: a reduction in the accuracy of identifying aligned cases. In other words, the new prompt enhances sensitivity to misalignment but increases the likelihood of incorrectly labeling aligned items as misaligned. Choosing the appropriate prompt, therefore, requires balancing these competing priorities, maximizing the detection of misalignments while minimizing false alarms for aligned cases. The optimal balance will depend on the specific goals and constraints of the classification task.

Overall, the results, summarized in Table S8 and Figures S1 and S2, indicate that incorporating the second prompt generally improved F1 scores across subjects and grades.

**Figure S1**

*Comparison of Accuracy Scores by Grade for Different Prompt and Alignment Types (GPT-4o-mini) for Math (Upper Panel) and Reading (Lower Panel).*

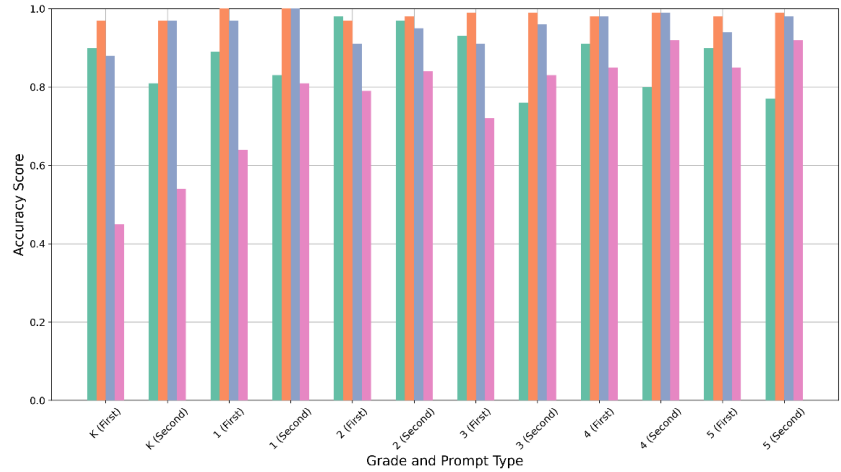

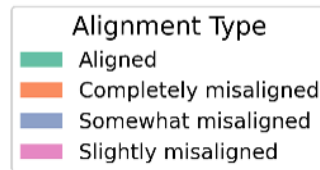

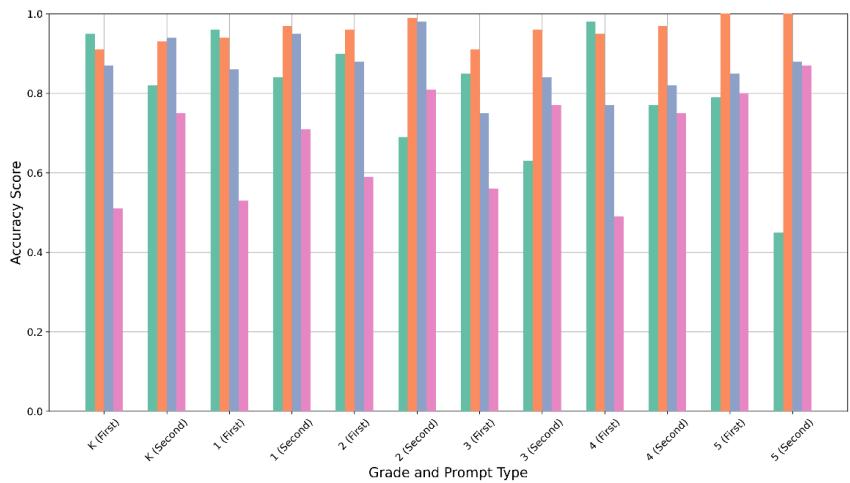

*Note*: The task is detecting whether item–skill pairs are aligned vs. misaligned. The second prompt improves sensitivity to misalignments but increases false positives for aligned cases.

**Table S8**

*Accuracy for Different Types of Prompts and Alignment.*

| Grade | Subject | Prompt type | Aligned | Completely misaligned | Somewhat misaligned | Slightly misaligned | F1 |
|---|---|---|---|---|---|---|---|
| K | Math | First | 0.90 | 0.97 | 0.88 | 0.45 | 0.85 |
| | | Second | 0.81 | 0.97 | 0.97 | 0.54 | 0.88 |
| K | Reading | First | 0.95 | 0.91 | 0.87 | 0.51 | 0.86 |
| | | Second | 0.82 | 0.93 | 0.94 | 0.75 | 0.90 |
| 1 | Math | First | 0.89 | 1.00 | 0.97 | 0.64 | 0.91 |
| | | Second | 0.83 | 1.00 | 1.00 | 0.81 | 0.94 |
| 1 | Reading | First | 0.96 | 0.94 | 0.86 | 0.53 | 0.87 |
| | | Second | 0.84 | 0.97 | 0.95 | 0.71 | 0.91 |
| 2 | Math | First | 0.98 | 0.97 | 0.91 | 0.79 | 0.94 |
| | | Second | 0.97 | 0.98 | 0.95 | 0.84 | 0.95 |
| 2 | Reading | First | 0.90 | 0.96 | 0.88 | 0.59 | 0.88 |
| | | Second | 0.69 | 0.99 | 0.98 | 0.81 | 0.91 |
| 3 | Math | First | 0.93 | 0.99 | 0.91 | 0.72 | 0.92 |
| | | Second | 0.76 | 0.99 | 0.96 | 0.83 | 0.92 |
| 3 | Reading | First | 0.85 | 0.91 | 0.75 | 0.56 | 0.83 |
| | | Second | 0.63 | 0.96 | 0.84 | 0.77 | 0.86 |
| 4 | Math | First | 0.91 | 0.98 | 0.98 | 0.85 | 0.95 |
| | | Second | 0.80 | 0.99 | 0.99 | 0.92 | 0.95 |
| 4 | Reading | First | 0.98 | 0.95 | 0.77 | 0.49 | 0.84 |
| | | Second | 0.77 | 0.97 | 0.82 | 0.75 | 0.88 |
| 5 | Math | First | 0.90 | 0.98 | 0.94 | 0.85 | 0.94 |
| | | Second | 0.77 | 0.99 | 0.98 | 0.92 | 0.94 |
| 5 | Reading | First | 0.79 | 1.00 | 0.85 | 0.80 | 0.90 |
| | | Second | 0.45 | 1.00 | 0.88 | 0.87 | 0.87 |

**Figure S2**

*Comparison of F1 Scores by Grade for Different Prompt Types (GPT-4o Mini). The upper panel represents math, while the lower panel represents reading.*

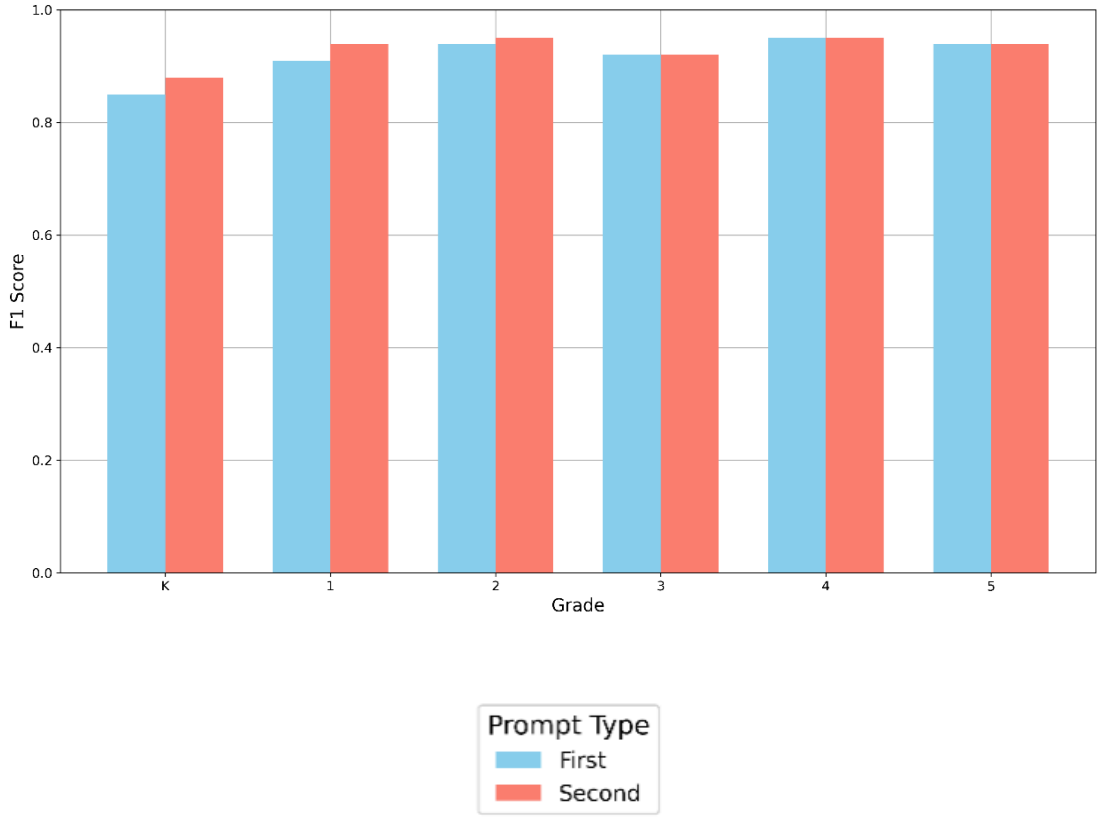

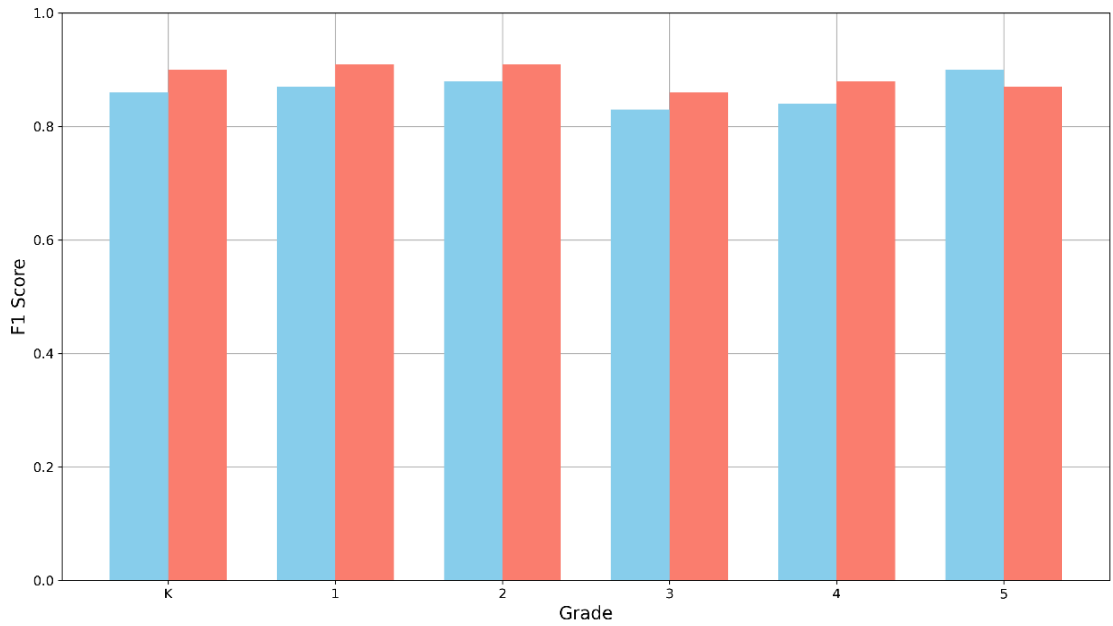

**Study 2**

An example prompt for the top-1 skill selection is shown below:

> You are provided with an educational item and a list of skills. Your task is to determine which one skill is most likely being measured by the item. Base your decision on clear and strong evidence. WARNING: You must return ONLY the Skill ID as your response. Example Output: 520.

**Table S9**

*Unique skill statements  by Grade and Subject*

| Grade Level | Number of Unique Skills | |
|:---:|:---:|:---:|
| | Math | Reading |
| K | 12 | 32 |
| 1 | 15 | 30 |
| 2 | 19 | 31 |
| 3 | 26 | 33 |
| 4 | 31 | 36 |
| 5 | 22 | 33 |

**Table S10**

*Accuracy of GPT-4o-mini in Matching Items to Skills Across Grades and Subjects*

| Subject | Grade | Items | Top 1 | Top 3 | Top 5 |
|---------|-------|-------|-------|-------|-------|
| **Math** | K | 64 | 79.68 | 85.94 | 90.62 |
| **Math** | 1 | 78 | 57.69 | 87.18 | 96.15 |
| **Math** | 2 | 103 | 80.58 | 92.23 | 94.17 |
| **Math** | 3 | 322 | 59.93 | 78.57 | 86.34 |
| **Math** | 4 | 596 | 70.13 | 84.40 | 87.08 |
| **Math** | 5 | 358 | 72.06 | 86.31 | 88.83 |
| **Reading** | K | 437 | 66.13 | 74.60 | 83.30 |
| **Reading** | 1 | 289 | 63.32 | 77.16 | 89.27 |
| **Reading** | 2 | 217 | 51.61 | 69.12 | 78.80 |
| **Reading** | 3 | 343 | 39.35 | 62.97 | 77.84 |
| **Reading** | 4 | 136 | 47.79 | 77.2 | 86.76 |
| **Reading** | 5 | 68 | 69.11 | 70.58 | 97.05 |

**Study 3**

**Table S11**

*Accuracy of Sentence Similarity in Skill Retrieval*

| Subject | Grade | Total Items | Top 1 | Top 3 | Top 5 | Top 10 | Top 15 | Top 20 |
|---------|-------|-------------|-------|-------|-------|--------|--------|--------|
| **Math** | K | 64 | 54.69 | 73.44 | 82.81 | 100.00 | 100.00 | 100.00 |
| **Math** | 1 | 78 | 34.62 | 66.67 | 84.62 | 100.00 | 100.00 | 100.00 |
| **Math** | 2 | 103 | 50.49 | 71.84 | 80.58 | 93.20 | 94.17 | 100.00 |
| **Math** | 3 | 322 | 41.61 | 63.98 | 73.60 | 87.27 | 90.68 | 92.24 |
| **Math** | 4 | 596 | 43.62 | 64.60 | 74.33 | 85.07 | 87.58 | 90.10 |
| **Math** | 5 | 358 | 36.31 | 62.57 | 75.14 | 86.59 | 88.83 | 91.34 |
| **Reading** | K | 437 | 37.53 | 62.70 | 75.29 | 87.19 | 94.28 | 95.65 |
| **Reading** | 1 | 289 | 35.29 | 52.25 | 63.32 | 82.35 | 90.31 | 93.77 |
| **Reading** | 2 | 217 | 42.86 | 64.06 | 75.58 | 86.18 | 91.71 | 94.47 |
| **Reading** | 3 | 343 | 19.24 | 35.28 | 49.56 | 67.93 | 81.63 | 88.92 |
| **Reading** | 4 | 136 | 32.35 | 61.03 | 68.38 | 74.26 | 82.35 | 88.24 |
| **Reading** | 5 | 68 | 8.82 | 42.65 | 58.82 | 75.00 | 86.76 | 92.65 |